\documentclass[runningheads]{llncs}
\PassOptionsToPackage{table}{xcolor}

\usepackage{eccv}

\usepackage{eccvabbrv}

\usepackage{graphicx}
\usepackage{booktabs}
\usepackage{multirow}
\usepackage{tabularx}
\usepackage{colortbl}
\usepackage{xcolor}
\usepackage{makecell}
\usepackage{wasysym}
\usepackage{wrapfig}

\newcommand{\gray}[1]{\cellcolor{gray!8}{#1}}
\usepackage[accsupp]{axessibility}  

\usepackage{hyperref}

\usepackage{orcidlink}

\newcommand{\dataset}{CMA-Loc\xspace}
\newcommand{\model}{GAGeo}

\begin{document}

\title{Beyond 2D Matching: A Unified Single-Stage Framework for Geometry-Aware Cross-View Object Geo-Localization} 

\titlerunning{Geometry-Aware Cross-View Object Geo-Localization}

\author{
Liyao Wang\inst{1}\textsuperscript{*} \and
Ruipu Wu\inst{1}\textsuperscript{*} \and
Haojun Xu\inst{1}\textsuperscript{*} \and
Lei Shi\inst{2} \\
Linjiang Huang\inst{1}\textsuperscript{\dag} \and
Si Liu\inst{1}
}

\authorrunning{L.~Wang et al.}

\institute{
Beihang University, Beijing, China \and
Meituan, Beijing, China \\
\email{\{bastien\_wu,xuhaojun123,ljhuang,liusi\}@buaa.edu.cn},
\email{shilei53@meituan.com}\\
\url{https://cipual.github.io/GAGeo-project-page/}
}

\maketitle

\begingroup
\renewcommand{\thefootnote}{*}
\footnotetext{Equal contribution.}
\renewcommand{\thefootnote}{\dag}
\footnotetext{Corresponding author.}
\endgroup

\begin{figure}
    \centering
    \includegraphics[width=1\linewidth]{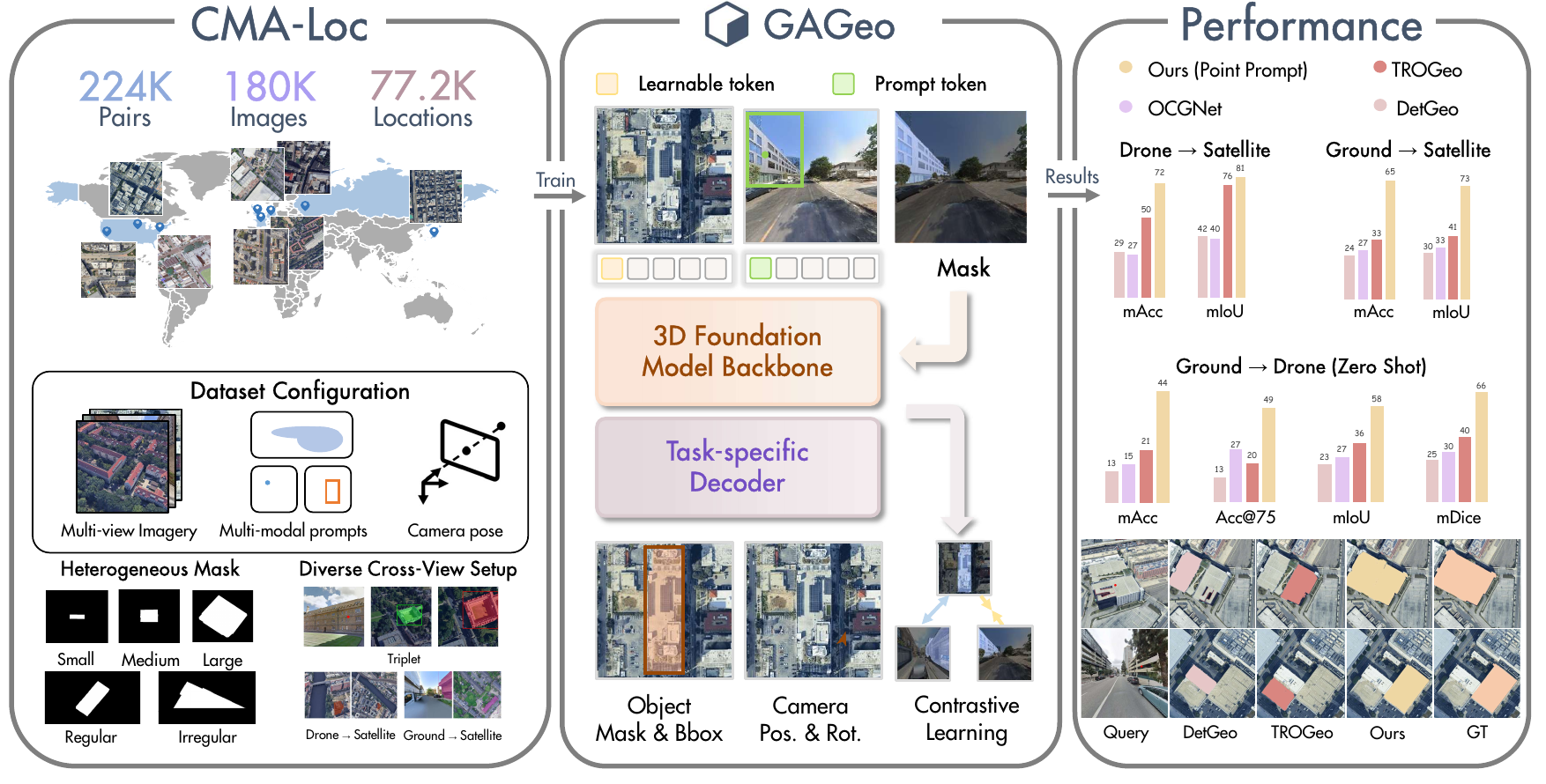}
    \caption{\textbf{Overview of \dataset dataset and \model ~framework.} We introduce \dataset, a large-scale building dataset for advancing cross-view geo-localization, featuring 224K instance pairs and 180K images across 77.2K locations. It incorporates multi-view imagery, multi-modal prompts, and camera pose to simulate complex real-world scenarios. Leveraging \dataset, we propose \model, a geometry-aware cross-view object geo-localization framework that adapts a 3D Foundation Model backbone with task-specific decoders in a single-stage manner, achieving state-of-the-art performance in drone-to-satellite, ground-to-satellite, and zero-shot ground-to-drone setups.}
    \label{fig:placeholder}
\end{figure}

\begin{abstract}
Cross-view object geo-localization (CVOGL) aims to locate a target object from a query view (e.g., ground or drone) within a geo-tagged reference image (e.g., satellite). Existing approaches heavily rely on 2D appearance matching and are constrained by limited datasets lacking geometric metadata, diverse prompts, and standard field-of-view imagery. To address these intertwined challenges, we first introduce \dataset, a large-scale, high-fidelity building dataset comprising over 220,000 ground-satellite and drone-satellite pairs. It provides multi-modal prompts (points, boxes, masks) and camera poses to enable flexible target referring and explicit spatial modeling. Furthermore, we propose a novel single-stage Geometry-Aware Geo-localization framework (\model), built upon the permutation-equivariant 3D foundation model $\pi^3$. By seamlessly integrating visual features, referring prompts, and learnable task tokens, our model adapts the inherited 3D prior to jointly predict bounding boxes, segmentation masks, and camera poses in a single forward pass. Additionally, we introduce a contrastive loss that utilizes the satellite view as a universal anchor, implicitly aligning ground and drone representations to enable zero-shot ground-to-drone localization without requiring triplet training data. Extensive experiments demonstrate that our approach significantly outperforms state-of-the-art methods, exhibiting exceptional generalization ability in unseen scenes and novel cross-view setups. 
\end{abstract}

\section{Introduction}
\label{sec:intro}
In real-world applications ranging from disaster monitoring~\cite{hansch2022spacenet, zhang2024good} to drone navigation~\cite{mithun2023cross}, there is a growing need to ground specific target objects onto geo-tagged reference imagery (e.g., satellite views), thereby decoupling spatial positioning from reliance on external signals. Inherently, this task requires building a spatial bridge that projects local, ego-centric observations onto a global coordinate system, seamlessly shifting the paradigm from recognizing ``what'' is in the scene to precisely localizing ``where'' it resides in the physical world.

Driven by this practical need, researchers have curated various cross-view datasets, evolving from coarse-grained image-level retrieval~\cite{CVUSA2017,CVACT2019, vigor2021, university2020, denseuav2023} to recent object-centric datasets~\cite{DetGeo, TROGeo}. However, current research is severely hindered by coupled limitations in both existing datasets and methodologies. On the data side, existing resources rely exclusively on panoramic images that introduce severe geometric distortions, failing to reflect the standard field-of-view (FoV) of practical cameras. Furthermore, they are restricted to simple point-coordinate prompts, lack geometric metadata (e.g., camera positions and rotations), and suffer from limited scale (merely 12,000 pairs for CVOGL~\cite{DetGeo}) and scene variability. 
Beyond these data deficiencies, prevailing research conventionally treats CVOGL as a 2D matching problem. Consequently, most existing frameworks~\cite{workman2015wide, hu2018cvm, shi2022beyond, SAFA, DetGeo, Sample4geo, TROGeo} rely on appearance-based cues, failing to fully exploit the inherent 3D geometric structure. Lacking such 3D awareness, these models struggle to bridge the spatial gap, leaving them vulnerable to perceptual ambiguity and visual inconsistencies, which limit their robust generalization.


To tackle these intertwined challenges, we propose a comprehensive framework comprising a high-fidelity building dataset, \textbf{\dataset}, and a novel Geometry-Aware GEO-localization approach \textbf{\model}. First, to bridge the data gap, \dataset is constructed via two tailored pipelines. To construct ground-satellite pairs, we collect imagery from Google Street View~\cite{goolgeStreetView} and Google Earth~\cite{gooleEarth}.
We then associate OpenStreetMap-derived~\cite{OSM} satellite footprints with SAM3-generated~\cite{SAM3} ground masks based on geometric cues like orientation and scale.
For drone-satellite pairs, we utilize a Cesium rendering pipeline to deterministically extract and pair cross-view masks via unique building IDs. Finally, to improve boundary quality across the dataset, all non-SAM-generated masks are refined using SAM3. From these refined masks, we derive bounding boxes and inner point prompts as additional referring prompts, followed by rule-based filtering to reduce low-quality or information-sparse cases.
Crucially, \dataset also provides accurate camera poses for each pair, unlocking the capability for explicit spatial modeling and geometric supervision.
Through this rigorous construction pipeline, we ultimately curate a large-scale dataset comprising $112,063$ ground-satellite and $111,704$ drone-satellite instance pairs across eight global cities, resolving the data scale and scene diversity constraints of prior works. Additionally, in \dataset's test set, we incorporate a manually annotated subset, which comprises ground-drone-satellite triplets from University-1652~\cite{university2020}, serving as a dedicated dataset to evaluate models' zero-shot and cross-domain generalization.

To overcome the limitations of pure 2D appearance matching, our methodology shifts the paradigm by harnessing the powerful geometric priors embedded in 3D foundation models~\cite{wang2025vggt, liu2025worldmirror, zhang2025emergent}. Specifically, we adopt $\pi^3$~\cite{wang2025pi} as our framework's backbone. A primary motivation for this choice is that $\pi^3$ processes input views in a permutation-equivariant manner; it establishes a shared representation space without anchoring to any initial frame's perspective, thereby reducing viewpoint biases.
Leveraging this property, we propose a unified single-stage framework capable of processing both ground-to-satellite and drone-to-satellite inputs. The model first fuses visual tokens from a frozen DINOv2~\cite{Dinov2} with high-fidelity prompt tokens from a SAM2-pretrained encoder. Unlike previous paradigms that rely on cumbersome two-stage networks for detection and segmentation (e.g., TROGeo~\cite{TROGeo}) or append heavy, isolated prediction heads to a backbone, our design is remarkably streamlined. We seamlessly integrate learnable tokens and prompt tokens as additional inputs for $\pi^3$, augmenting its alternating local and global attention layers with a specialized masking scheme designed for multi-task decoding. Through a single forward pass, this single-stage structure yields diverse multi-task predictions, including bounding boxes, segmentation masks, camera position and rotation.

Furthermore, since collecting perfectly aligned ground-drone-satellite triplets in the real world is notoriously difficult, we introduce a contrastive learning objective optimized purely on available pairs. Conceptually akin to ImageBind~\cite{girdhar2023imagebind}, which binds diverse modalities to a shared latent space using images as the central anchor, we utilize the satellite view as a universal intermediate bridge. Benefiting from the view-agnostic nature of $\pi^3$, aligning mask-pooled object representations within ground-satellite and drone-satellite pairs implicitly pulls the ground and drone representations closer together. This strategy circumvents the reliance on scarce triplet training data, unlocking the capability for zero-shot ground-to-drone geo-localization.

Extensive experiments demonstrate that our framework sets a new state-of-the-art in the cross-view object geo-localization task, significantly outperforming existing approaches. Notably, in the ground-to-satellite setup, our method achieves a significant improvement of 34.38\% mAcc on the object detection task and 33.36\% mIoU on the object segmentation task. Meanwhile, it also exhibits robust generalization to unseen dataset and novel ground-to-drone setup, evidently outperforming previous best-performing methods. Furthermore, comprehensive ablation studies validate the effectiveness of each proposed strategy.

\section{Related Works}
\label{sec:related_work}

\textbf{Cross-View Image Geo-Localization.}
Early cross-view geo-localization formulated the task as an image retrieval problem~\cite{lin2013cross, zhang2024geodtr+}. Despite the availability of diverse multi-view datasets~\cite{CVUSA2017,CVACT2019,vigor2021,university2020,denseuav2023}, existing methods predominantly rely on holistic, 2D appearance-based metric learning~\cite{workman2015wide, hu2018cvm, shi2022beyond}. To address extreme viewpoint discrepancies, subsequent works incorporated geometric transformations~\cite{SAFA, video2bev} and spatial partitioning~\cite{wang2021each, dai2021transformer}. However, these approaches remain constrained by rigid geometric priors and lack a robust 3D prior, fundamentally limiting their spatial perception in complex environments.

\noindent \textbf{Cross-View Object Geo-Localization.}
Shifting from image-level retrieval, cross-view object geo-localization (CVOGL) targets specific objects using point prompts. Current methods typically frame this as cross-view detection or segmentation~\cite{DetGeo, TROGeo, zhu2025improving}, relying on complex multi-stage architectures~\cite{TROGeo,ReCOT} or perspective-specific algorithms~\cite{VAGeo, OCGNet}. Fundamentally confined to 2D feature alignment, these methods struggle with extreme viewpoint variations and repetitive visual patterns. In contrast, our work proposes a unified framework that processes multi-modal prompts and generates multi-modal outputs in a single forward pass, overcoming the structural bottlenecks of pure 2D alignment.

\noindent \textbf{Bridging Extreme Views via 3D Geometric Foundation Models.}
Framing cross-view grounding as a 3D problem is highly effective, as corresponding pixels across perspectives inherently project to the same 3D point~\cite{mast3r}. Recently, feed-forward 3D geometric foundation models~\cite{wang2025vggt,liu2025worldmirror,wang2025pi} have demonstrated an emergent understanding of extreme-view geometry~\cite{zhang2025emergent}, providing a robust scaffold for complex scenarios. Building on this observation, we introduce \(\pi^3\)~\cite{wang2025pi} to the CVOGL task. This novel geometry-aware paradigm distills intrinsic 3D topologies that remain invariant across heterogeneous perspectives, fundamentally resolving extreme view discrepancies and visual ambiguities.

\section{Dataset}
\label{sec:dataset}

In this section, we introduce \textbf{\dataset} (\textbf{C}ross-view \textbf{M}ulti-prompt \textbf{A}nnotated \textbf{Loc}alization dataset), a building-focused dataset designed to advance cross-view geo-localization by providing large-scale, multi-prompt geometric supervision. Spanning 77,200 geographic locations across 8 diverse cities (e.g., Tokyo, London, etc.), the dataset provides 112,063 ground-satellite and 111,704 drone-satellite instance pairs. By spanning diverse architectural styles and cultural environments, \dataset significantly exceeds the scale of existing datasets. The following sections detail the data collection process, the annotation pipeline, and dataset statistics.

\begin{figure}[!t]
    \centering
    \includegraphics[width=0.99\linewidth]{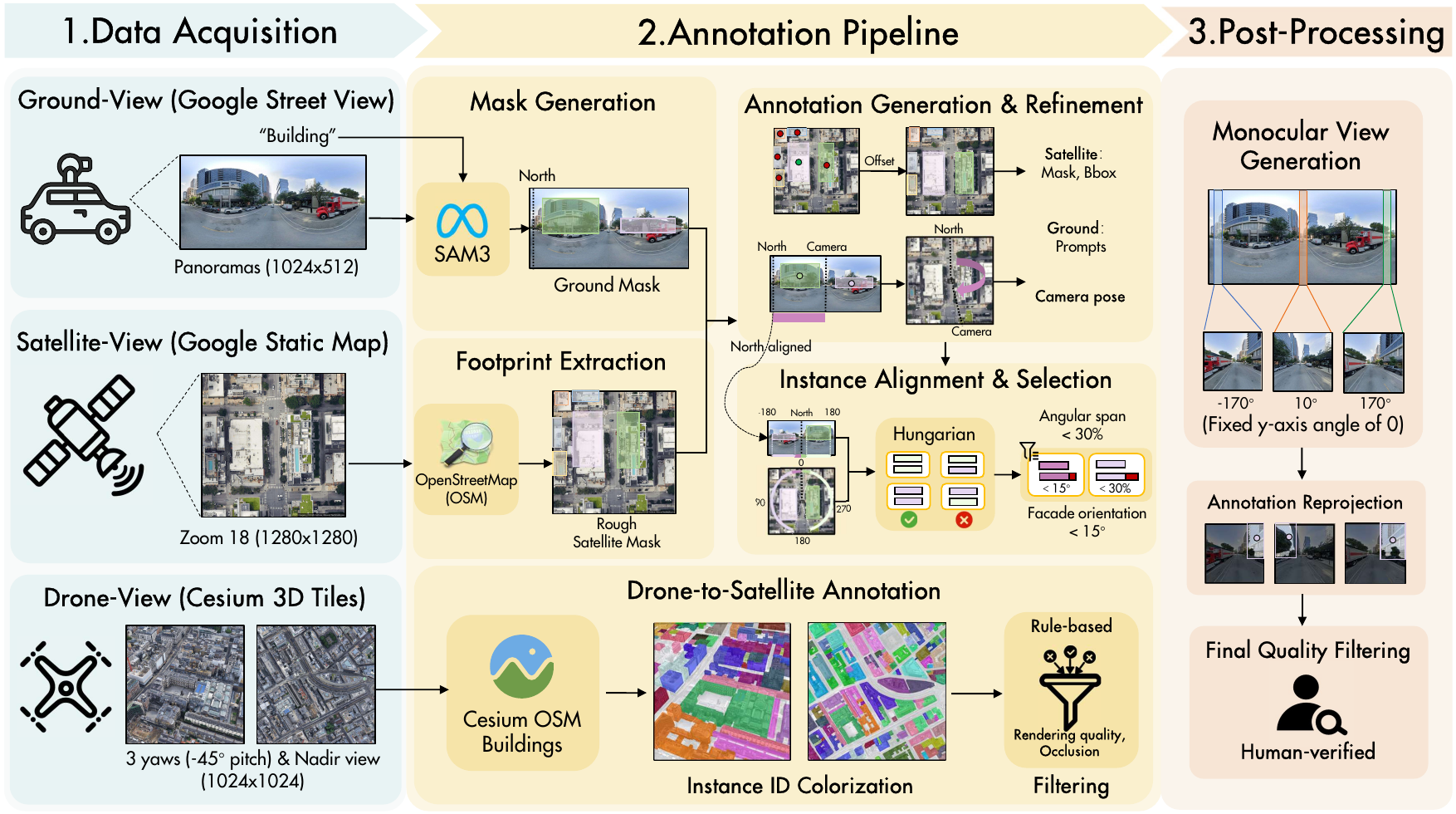}
    \caption{Illustration of the pipeline construction process for \dataset. We present two specialized workflows for generating \textbf{ground-to-satellite} and \textbf{drone-to-satellite} instance pairs for cross-view geo-localization. The process integrates Google Street View panoramas with Cesium-synthesized drone views and nadir satellite imagery, ensuring precise geometric alignment across diverse cross-view perspectives.}
    \label{fig:dataset}
\end{figure}

\subsection{Data Collection}
The data collection process of \dataset follows a query-reference paradigm, matching ground- and drone-view query images against universal satellite reference images via the pipeline shown in Fig.~\ref{fig:dataset}.


\noindent \textbf{Ground-to-Satellite Image Collection.} For the ground-view queries, following the protocol established by Omnicity~\cite{li2023omnicity}, we collect Google Street View panoramas ($1024 \times 512$) sampled at 65m intervals. To ensure geometric alignment, we record the GPS coordinates, panorama IDs, and north-aligned headings for each site. Corresponding satellite imagery is sourced from Google Earth at zoom level 18, providing $1280 \times 1280$ reference images.

\noindent \textbf{Drone-to-Satellite Image Synthesis.} Drone-to-satellite data is synthesized via a Cesium-based rendering pipeline, leveraging Google Photorealistic 3D Tiles for urban reconstruction. We sample these environments at a 0.005 spatial grid interval. At each location, we simulate three drone views with a fixed $-45^\circ$ pitch and yaw angles spaced at $120^\circ$ intervals with a random initial offset. Corresponding satellite references are generated via nadir orthographic rendering ($-90^\circ$ pitch) at the same coordinates. All images are rendered at $1024 \times 1024$ resolution, with camera Euler angles and grid coordinates stored as metadata.

\noindent \textbf{Ground-Drone-Satellite Image Triplet Preparation.} To validate models' zero-shot geo-localization capability in unseen cross-view setups and environments, we carefully select 1245 triplets from the University-1652 dataset~\cite{university2020} to prepare raw ground-drone-satellite image triplets. 

\subsection{Data Annotation}

Fig.~\ref{fig:dataset} depicts our automated annotation pipeline, which generates multi-modal prompts and geometric supervision without prohibitive manual labor.

\noindent \textbf{Ground-to-Satellite Annotation.} The ground-to-satellite annotation pipeline proceeds in three streamlined stages as follows:

1) Mask Generation and Refinement. We first extract initial building masks from ground panoramas using SAM3~\cite{SAM3}, filtering them by object size and occlusion. For satellite imagery, we generate high-fidelity masks by prompting SAM with OpenStreetMap (OSM)~\cite{OSM} footprints: using each footprint's center as a positive prompt and the five nearest neighbors as negative prompts. After correcting inherent OSM-SAM misalignments via global offset estimation, the masks are filtered by area, confidence, and IoU to remove unreliable or low-information cases. Finally, bboxes and interior points are extracted from these refined masks to serve as multi-modal prompts.

2) Geometric Instance Alignment. Next, we establish cross-view correspondences using spatial metadata. By leveraging north-aligned rotation angles, we compute each building's facade orientation and angular span. A ground-satellite pair is retained only if the facade orientation difference is less than \(15^\circ\) and the angular span discrepancy is within 30\%, which filters cases where occlusion or limited visible facade length makes the correspondence ambiguous. Additionally, camera pixel coordinates and north-relative orientations are explicitly recorded in the satellite images to provide rich pose annotations.

3) Monocular View Synthesis. Finally, to align with real-world deployment, we transform panoramas into standard monocular views.

\noindent \textbf{Drone-to-Satellite Annotation.} For the synthesized drone-to-satellite data, the annotation process is deterministic. We leverage Cesium OSM Buildings to generate instance masks via unique RGB encoding, ensuring exact cross-view correspondence and avoiding heuristic matching. Finally, to ensure dataset quality, image pairs are filtered to remove poor renderings, invalid masks, or significant occlusions (e.g., dense trees).

\noindent \textbf{Ground-Drone-Satellite Triplet Annotation.} \label{sec:dataset_annotation}
We construct a specialized evaluation subset by annotating the ground-drone-satellite triplets. Concretely, Initial point and bounding box prompts are first generated using Gemini~\cite{gemini2023}, which are then subjected to rigorous manual verification to ensure their fidelity. Finally, high-fidelity segmentation masks are produced using SAM3~\cite{SAM3}, utilizing the verified bounding boxes as prompts.

\begin{table}[t]
  \centering
  \caption{Comparison of cross-view localization datasets. G, D, S, Pt. denote ground, drone, satellite, and point, respectively. Mono. Ground View refers to standard FOV.}
  \label{tab:dataset_comparison}
  \renewcommand{\arraystretch}{1.1}
  \setlength{\tabcolsep}{4pt}
  \resizebox{\linewidth}{!}{%
  \begin{tabular}{@{}lcccccc@{}}
    \toprule
     & CVUSA & VIGOR & Univ-1652 & CVOGL & CVOGL-Seg
     & \gray{\textbf{\dataset}} \\
    \midrule
    \# Instance Pairs
      & --     & --      & --     & 12,478  & 12,478
      & \gray{223,767} \\
    \# Satellite Images
      & 44,416 & 105,214 & 951    & 5,836   & 5,836
      & \gray{77,200}  \\
    \# Ground Images
      & 44,416 & 90,618  & 2,921  & 5,279   & 5,279
      & \gray{93,144}  \\
    \# Drone Images
      & --     & --      & 51,355 & 5,279 & 5,279
      & \gray{10,086}  \\
    \midrule
    Cross-View Setup
      & G${\to}$S & G${\to}$S
      & \makecell{%
          \begin{tikzpicture}[scale=0.45, baseline=-2pt,
              every node/.style={font=\footnotesize, inner sep=1pt}]
            \node (G) at (0,{sqrt(3)*0.75})  {G};
            \node (D) at (-0.75,0) {D};
            \node (S) at ( 0.75,0) {S};
            \draw[<->,>=stealth, line width=.4pt] (G)--(D);
            \draw[<->,>=stealth, line width=.4pt] (D)--(S);
            \draw[<->,>=stealth, line width=.4pt] (S)--(G);
          \end{tikzpicture}}
      & G/D${\to}$S & G/D${\to}$S
      & \gray{G/D${\to}$S} \\
    Annotation Grain
      & Image & Image & Image & Object & Mask
      & \gray{Mask} \\
    Prompt Type
      & -- & -- & -- & Point & Point
      & \gray{Pt.\,/\,Bbox\,/\,Mask} \\
    \midrule
    Mono. Ground View
      & \Circle & \Circle & \CIRCLE & \Circle & \Circle
      & \gray{\CIRCLE} \\
    Orientation Info
      & \CIRCLE & \CIRCLE & \Circle & \Circle & \Circle
      & \gray{\CIRCLE} \\
    Triplet Eval
      & \Circle & \Circle & \CIRCLE & \Circle & \Circle
      & \gray{\LEFTcircle} \\
    \bottomrule
  \end{tabular}}
\end{table}

\subsection{Dataset Statistics}
\label{sec:dataset_statistics}

Both the drone-to-satellite and ground-to-satellite tasks localize targets via cross-view correspondence, inherently sharing a unified problem formulation. Unlike CVOGL, which addresses these tasks in isolation, we introduce a single, comprehensive dataset to investigate both paradigms jointly. As summarized in Tab.~\ref{tab:dataset_comparison}, \dataset offers several distinct advantages over existing datasets:

First, it features a much larger data scale, being approximately 18\(\times\) larger than previous datasets. Second, it supports rich, multi-modal prompts, including points, bounding boxes, and masks, overcoming the limitations of datasets like CVOGL~\cite{DetGeo} that rely solely on sparse point prompts. Third, it provides geometric annotations, including precise observer positions and camera poses, delivering the explicit supervision necessary for spatial understanding. 

Beyond scale and annotation richness, \dataset spans 8 globally distributed cities with highly distinct architectural styles. This geographic diversity introduces substantial variations in object scale, shape, and occlusion, thereby significantly elevating the challenge of robust detection and segmentation. Furthermore, to better reflect real-world deployment conditions, \dataset discards impractical panoramic ground images in favor of standard monocular views, enhancing the benchmark's practical applicability.

Finally, as detailed in Sec.~\ref{sec:dataset_annotation}, the evaluation protocol of \dataset goes beyond standard in-domain testing. Its evaluation set contains not only ground-satellite and drone-satellite pairs aligned with the training distribution, but also perfectly aligned ground-drone-satellite triplets meticulously curated to validate the models' cross-view generalization capabilities.



\section{Method}
\label{sec:method}

In this section, we first formally define the task formulation for cross-view object grounding and localization. Then, Section \ref{subsec:preliminary} briefly reviews the preliminary architecture of contemporary 3D Foundation Models (3DFMs). Building upon this foundation, Section \ref{subsec:pipeline} details our proposed \textbf{G}eometry-\textbf{A}ware Cross-View \textbf{Geo}-Localization framework (\textbf{\model}), which seamlessly integrates multi-modal geometric prompts and task-specific tokens into a unified, single-stage transformer framework. Finally, Sec.~\ref{subsec:optimization} presents the joint multi-task optimization objectives employed to train the entire framework end-to-end.

\noindent \textbf{Task Formulation.} Given a query image $I_q$ (e.g. from ground or drone) and a reference image $I_r$ (e.g. from satellite), our objective is to learn a mapping function $\mathcal{F}(I_q, I_r, p) \rightarrow y_r$ that identifies and locates the target object in $I_r$ based on a spatial referring prompt $p$ defined in $I_q$. In our framework, the referring prompt $p$ is a multi-modal geometric prompt that can be represented as a point $\mathbf{p}_q = (x, y)$, a bounding box $\mathbf{b}_q = (x, y, w, h)$, or a binary mask $\mathbf{m}_q \in \{0, 1\}^{H \times W}$ that indicate the exact location of the target object in $I_q$. The model processes the cross-view visual context to output the target’s state $y_r$ in the reference domain, which is manifested as a predicted bounding box $\hat{\mathbf{b}}_r$ for object-level localization or a predicted mask $\hat{\mathbf{m}}_r$ for pixel-level segmentation. 

\begin{figure}[!t]
    \centering
    \includegraphics[width=0.99\linewidth]{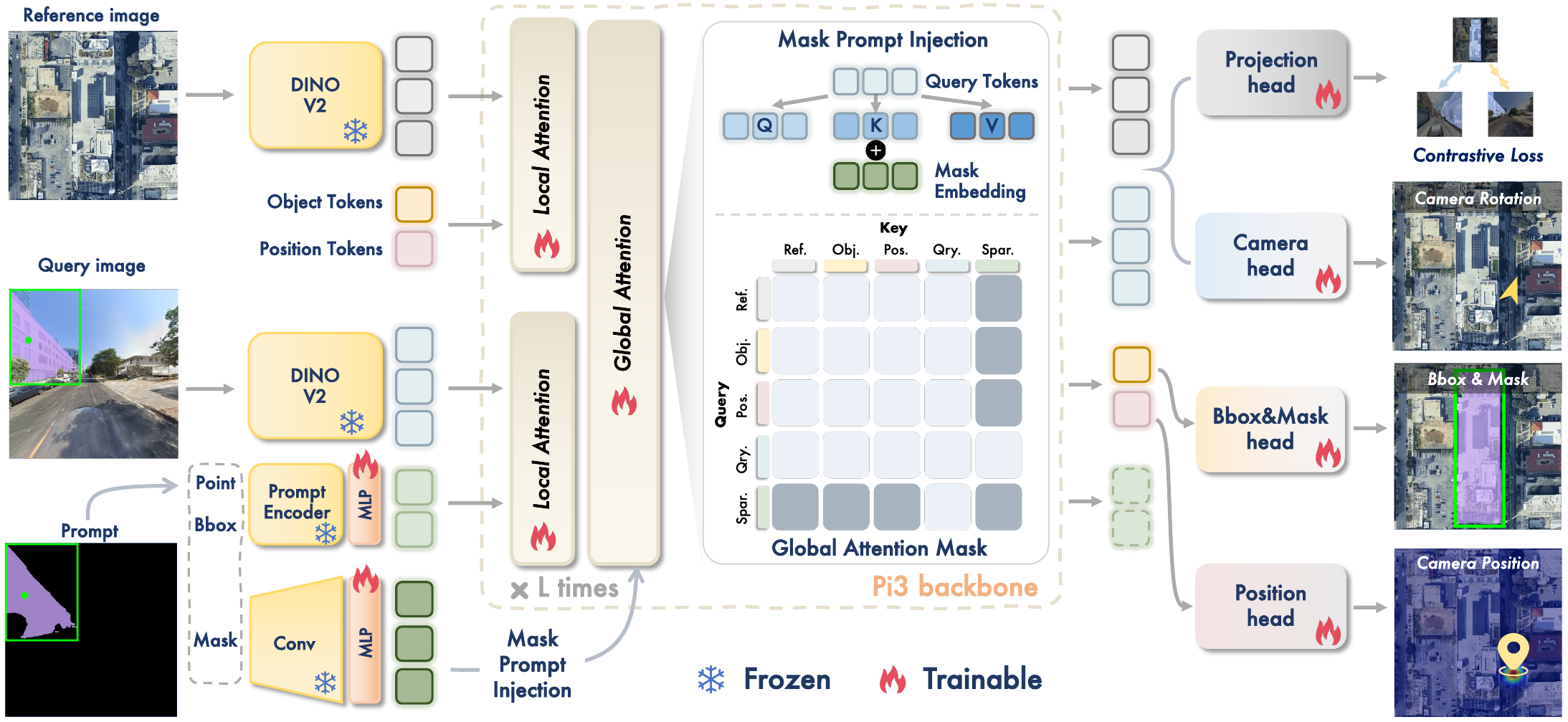}
    \caption{Overview of \model, which integrates multi-modal geometric prompts and task-specific tokens into a unified, single-stage, multi-task transformer framework.}
    \label{fig:method}
\end{figure}

\subsection{Preliminary}
\label{subsec:preliminary}

\noindent \textbf{3DFM Architecture.} Contemporary 3DFMs~\cite{wang2025vggt, wang2025pi, liu2025worldmirror} typically share a unified architectural framework. An encoder \(\varepsilon\) first maps input images into patch-level embeddings. Specifically, given two input images, \(I_1\) and \(I_2\), each divided into \(N_p = H_p \times W_p\) patches, the tokens are processed by a shared transformer backbone that interleaves \emph{local} and \emph{global} attention blocks. 
Local attention blocks process each view independently:
$
\mathbf{T}_{\text{frame}}^{(i)} \in \mathbb{R}^{N_p \times D}$ where $i \in \{1,2\}.
$
Conversely, global attention blocks facilitate cross-view interaction by aggregating tokens from both images via concatenation:
\begin{equation}
\mathbf{T}_{\text{global}} = [\mathbf{T}_{\text{frame}}^{(1)}, \mathbf{T}_{\text{frame}}^{(2)}] \in \mathbb{R}^{2 N_p \times D}.
\end{equation}
Both variants employ standard self-attention. For layer \(l\) and head \(h\), queries $\mathbf{Q}=f_Q(\mathbf{T}_*)$, keys $\mathbf{K}=f_K(\mathbf{T}_*)$, and values $\mathbf{V}=f_V(\mathbf{T}_*)$ are derived via learned linear projections, yielding the attention weights:
\begin{equation}
\mathbf{A}_h^{(l)} = \text{softmax}(\mathbf{Q} \mathbf{K}^T / \sqrt{d_h}),
\end{equation}
where \(d_h\) is the head dimensionality. Finally, the processed tokens are fed into task-specific decoding heads, such as a camera head for pose estimation and a position head for localization.

\subsection{Geometry-Aware CVOGL Pipeline} \label{subsec:pipeline}
 Extending the \(\pi^3\) framework~\cite{wang2025pi}, our pipeline augments the input sequence with task-specific tokens and referring prompt tokens tailored for CVOGL, rather than treating the query and reference images as standard stereo pairs. Critically, while conventional 3DFMs anchor their spatial representations to the first frame, our approach leverages the permutation-equivariant design of \(\pi^3\)~\cite{wang2025pi}. By deliberately discarding order-dependent components, such as frame-wise positional embeddings and camera tokens, the model reduces view-order inductive biases. This inherited permutation equivariance is important for generalization across unseen cross-view setups (e.g., ground-to-drone). More importantly, it enables the model to support joint training across diverse cross-view setups within a unified framework, effectively leveraging heterogeneous cross-view pairs.

\noindent \textbf{Image and Prompt Encoding.} We employ a frozen DINOv2 encoder to extract patch-level tokens \(T_r, T_q \in \mathbb{R}^{N_p \times D}\) from both views. Concurrently, referring prompts are processed using a pretrained SAM2 module, which generates two types of representations: a prompt encoder produces sparse tokens \(T_p \in \mathbb{R}^{N_e \times D}\) (\(N_e=1\) for points and \(N_e=2\) for box corners), while a CNN encoder extracts dense embeddings \(E_d \in \mathbb{R}^{N_p \times D}\) for mask prompts.


\noindent \textbf{Prompt Injection and Cross-View Interaction.} 
During this stage, we leverage the inherent advantages of the \(\pi^3\) backbone. Specifically, its interleaved local and global attention mechanisms provide geometry-aware representation capabilities and cross-view information exchange. This architectural design naturally allows us to seamlessly integrate task-specific learnable tokens and referring prompts directly into the sequence processing. Consequently, our framework forms a fully transformer-based single-stage paradigm~\cite{songvidt, fang2021you}, unifying feature extraction, cross-view interaction, and task prediction within a single forward pass. Compared to previous cross-view methods that rely on cumbersome two-stage networks for detection and segmentation (e.g., TROGeo~\cite{TROGeo}) or append isolated prediction decoders to the backbone, our streamlined approach is simpler and more efficient.

Concretely, we employ an asymmetric token injection strategy tailored for our task formulation. Since the referring prompt is defined on the query image, while the target predictions (e.g., detection, segmentation) are executed on the reference image, we decouple their token assignments. Specifically, for the reference stream, we introduce learnable task tokens \(T_l \in \mathbb{R}^{(N_{obj} + N_{pos}) \times D}\), comprising object tokens \(T_{obj}\) and position tokens \(T_{pos}\), which are concatenated with the reference tokens to yield \(\overline{T}_r = [T_r, T_l]\) for generating the task results. 

For the query stream, prompt integration employs a modality-dependent strategy. Sparse prompt tokens $T_p$ are directly concatenated with the query tokens to form $\overline{T}_q = [T_q, T_p]$. In contrast, for dense mask embeddings $E_d$, we diverge from the standard SAM-style approach~\cite{sam2}, which adds these embeddings directly to the input image tokens. To ensure better generalization and preserve the symmetric input structure expected by the $\pi^3$ backbone, we utilize $E_d$ strictly as a modulation signal within the global attention blocks. Specifically, during global attention, $E_d$ is added element-wise exclusively to the \emph{keys} ($\mathbf{K}$) of the query image tokens. This key-side modulation avoids directly altering the layer's output feature distribution; instead, it refines the attention weights, effectively guiding the reference tokens to aggregate features from the mask-related regions of the query image during global interaction.

Within global attention blocks, we regulate cross-view interaction via a directional isolation mask. Specifically, sparse prompt tokens $T_p$ can attend to query tokens $T_q$ to absorb spatial priors, but are strictly isolated from the reference stream ($T_r$ and $T_l$), and vice versa. This isolation is essential: since geometric prompts are inherently query-view specific, exposing them to reference tokens provides no contextual benefit and risks corrupting reference features.

Through this stage, we can obtain the updated tokens $\hat{T}_q$, $\hat{T}_r$, $\hat{T}_l$, corresponding to the query features, the reference features and the task tokens, respectively.

\noindent \textbf{Lightweight Task-Specific Decoding.} 
The output task tokens $\hat{T}_l$ are partitioned into object tokens $\hat{T}_{obj}$ and position tokens $\hat{T}_{pos}$ for lightweight decoding. For object prediction, $\hat{T}_{obj}$ is processed by two parallel heads: a DETR-style~\cite{DETR} 3-layer MLP $\mathcal{D}_{det}$ for bbox regression, and a SAM-style~\cite{SAM} hyper-network $\mathcal{D}_{seg}$ that dynamically generates convolution kernels from $\hat{T}_{obj}$, which are applied to bilinearly upsampled reference features $\hat{T}_r^{\uparrow}$ for mask prediction. For the estimation of the observer's position (camera position) in the query image, a position head $\mathcal{D}_{pos}$ computes a spatial heatmap $H$ via a dot product between $\hat{T}_{pos}$ and upsampled reference features $\hat{T}_r^{\uparrow}$. Finally, following $\pi^3$~\cite{wang2025pi}, we use a camera head $\mathcal{D}_{cam}$ to process the updated tokens $\hat{T}_q$ and $\hat{T}_r$ to estimate the orientation of the query camera relative to the reference coordinate frame.


\subsection{Optimization Objective} \label{subsec:optimization}
The proposed framework is trained end-to-end using a joint multi-task objective. To ensure robust convergence and training stability, we apply deep supervision across intermediate decoder layers \(\mathcal{K} = \{4, 11, 17\}\). The overall loss function is formulated as follows, where $w_k$ is the weight for the $k$-th layer:
\begin{equation}
    \mathcal{L}_{total} = \sum_{k \in \mathcal{K}} w_k \left( \mathcal{L}_{grd}^{(k)} + \mathcal{L}_{pos}^{(k)} + \mathcal{L}_{rot}^{(k)} \right) + \mathcal{L}_{cl}.
\end{equation}
For more details, please refer to the supplementary material.

\noindent \textbf{Grounding Loss (\(\mathcal{L}_{grd}\)).} We employ Hungarian matching to assign the optimal prediction to the ground truth for object tokens \(T_{obj}\). The matched token is supervised via:
\begin{equation}
\mathcal{L}_{grd} = \lambda_{cls} \mathcal{L}_{focal} + \lambda_{box} (\mathcal{L}_{L1} + \mathcal{L}_{giou}) + \lambda_{mask} (\mathcal{L}_{bce} + \mathcal{L}_{dice}),
\end{equation}
which combines a binary focal loss for classification, \(L_1\) and GIoU losses for bounding box regression, and BCE and Dice losses for mask generation.

\noindent \textbf{Camera Pose Loss (\(\mathcal{L}_{pos} + \mathcal{L}_{rot}\)).} 
The query camera's pixel position (\(\mathcal{L}_{pos}\)) is supervised using a focal loss variant~\cite{cornernet} between the predicted heatmap and a Gaussian-augmented ground truth heatmap. For camera's orientation (\(\mathcal{L}_{rot}\)), we minimize the geodesic error on the \(\mathrm{SO}(3)\) manifold:
\begin{equation}
   \mathcal{L}_{rot} = \arccos\left(tr(\hat{R}^T R_{gt}) / 2 - 0.5\right),
\end{equation}
where \(\hat{R}\) and \(R_{gt}\) denote the predicted and GT rotation matrices, respectively.

\noindent \textbf{Contrastive Learning Loss (\(\mathcal{L}_{cl}\)).} To bridge the cross-view representation gap, we utilize a contrastive loss~\cite{moco} to align mask-pooled object features, using the satellite view as an anchor. We first project \(\hat{T}_q\) and \(\hat{T}_r\), then average pool them via the target object masks to obtain \(z_r\) and \(z_q\). The paired query embedding is treated as the positive \(z_q^+\), while other query embeddings in the MoCo queue form negatives \(z_q^-\):
\begin{equation}
\mathcal{L}_{cl} = -\log \frac{\exp(z_r \cdot z_q^+ / \tau)}{\exp(z_r \cdot z_q^+ / \tau) + \sum_{z_q^- \in Q} \exp(z_r \cdot z_q^- / \tau)},
\end{equation}
where \(z_r\) is the reference-view anchor, \(Q\) denotes the negative sample queue, and \(\tau=0.07\).

\section{Experiments}
\label{sec:experiments}

To demonstrate the superiority of \model, we first detail our experimental setup in Sec.~\ref{subsec:experimental_setup}. Then, we present a comprehensive comparison with state-of-the-art approaches across multiple CVOGL tasks in Sec.~\ref{subsec:comparisons} along with detailed ablation studies in Sec.~\ref{subsec:ablations}. Finally, we present qualitative comparison results in Sec.~\ref{subsec:qualitative_results}.


\subsection{Experimental Setup}
\label{subsec:experimental_setup}

\noindent \textbf{Datasets.}
Departing from isolated training protocols, \emph{all trainable compared methods are jointly trained across ground/drone-to-satellite setups in \dataset} following their original implementations with dataset-specific anchor or scale adaptation. For evaluation, we use the \dataset test set, containing 4,197 seen and 4,220 unseen instance pairs. Zero-shot ground-to-drone generalization is assessed using 1,245 annotated ground-drone-satellite triplets (see Sec.~\ref{sec:dataset_annotation}). Furthermore, we evaluate out-of-distribution scene generalization for object detection and segmentation on CVOGL~\cite{DetGeo} and CVOGL-Seg~\cite{TROGeo} across ground/drone-to-satellite setups.


\noindent \textbf{Evaluation Metrics.}
We use Acc@K (K $\in \{75\%, 50\%\}$) and mAcc (the average accuracy over IoU thresholds from 0.5 to 0.95 with a 0.05 interval) to evaluate the cross-view object detection task. For the cross-view object segmentation task, we use mIoU, mDice, AAE, and ME (see the supplementary material for details).

\noindent \textbf{Implementation Details.}
During training, the DINOv2 encoder and the SAM prompt encoder are kept frozen. Meanwhile, the $\pi^3$ backbone, together with the camera head, are initialized from pretrained weights. All other components are trained from scratch. (see further details in the supplementary material).

\subsection{Main Comparisons}
\label{subsec:comparisons}
As shown in Table~\ref{tab:cross_view} and \ref{tab:seg_comparison}, \model consistently surpasses existing SOTA methods across all metrics on the \dataset test set, irrespective of the prompt modality. Notably, this superiority is achieved as \emph{all trainable compared methods are fully trained on our dataset}. Furthermore, our approach effectively bridges the performance gap between seen and unseen environments. Remarkably, \model yields higher segmentation accuracy in a single forward pass than previous methods relying on an additional SAM Prompt Stage (SPS)~\cite{TROGeo}, although its inference cost is still dominated by the 3DFM backbone. This robust performance extends to the zero-shot setting on CVOGL-Seg (Tab.~\ref{tab:zero_shot_cvogl}), where our method nearly doubles the mAcc and mIoU of the prior SOTA without any domain-specific fine-tuning, thereby highlighting its exceptional transferability. Finally, as shown in Tab.~\ref{tab:zero_shot_g2d}, \model also outperforms the SPS-enhanced baseline in the unseen ground-to-drone setup, validating its generalization to novel viewpoint setup.

\begin{table*}[!t]
\centering
\caption{\textbf{Comparison on cross-view object detection task.} The best and second-best results are marked in \textbf{bold} and \underline{underline}, respectively.}
\label{tab:cross_view}
\setlength{\tabcolsep}{2pt}
\resizebox{\textwidth}{!}{%
\begin{tabular}{@{} l *{3}{c} @{\hskip 6pt} *{3}{c} @{\hskip 8pt} *{3}{c} @{\hskip 6pt} *{3}{c} @{}}
\toprule
\multirow{3}{*}{\textbf{Method}}
  & \multicolumn{6}{c}{\textbf{Drone} $\rightarrow$ \textbf{Satellite}}
  & \multicolumn{6}{c}{\textbf{Ground} $\rightarrow$ \textbf{Satellite}} \\
\cmidrule(lr){2-7} \cmidrule(lr){8-13}
  & \multicolumn{3}{c}{Seen} & \multicolumn{3}{c}{Unseen}
  & \multicolumn{3}{c}{Seen} & \multicolumn{3}{c}{Unseen} \\
\cmidrule(lr){2-4} \cmidrule(lr){5-7} \cmidrule(lr){8-10} \cmidrule(lr){11-13}
  & mAcc${\uparrow}$ & Acc@75${\uparrow}$ & Acc@50${\uparrow}$
  & mAcc${\uparrow}$ & Acc@75${\uparrow}$ & Acc@50${\uparrow}$
  & mAcc${\uparrow}$ & Acc@75${\uparrow}$ & Acc@50${\uparrow}$
  & mAcc${\uparrow}$ & Acc@75${\uparrow}$ & Acc@50${\uparrow}$ \\
\midrule
RK-Net~\cite{RK-Net}
  & 7.47 & 1.97 & 27.66
  & 7.02 & 1.44 & 28.36
  & 0.11 & 0.05 & 0.44
  & 0.15 & 0.00 & 0.73 \\
L2LTR~\cite{L2LTR }
  & 5.58 & 1.42 & 20.47
  & 4.07 & 0.89 & 16.66
  & 0.20 & 0.05 & 0.93
  & 0.19 & 0.00 & 1.15 \\
TransGeo~\cite{TransGeo }
  & 8.71 & 2.29 & 32.05
  & 7.15 & 1.44 & 29.35
  & 0.33 & 0.10 & 1.77
  & 0.13 & 0.00 & 0.78 \\
SAFA~\cite{SAFA}
  & 6.90 & 1.69 & 25.82
  & 6.03 & 1.19 & 24.44
  & 0.40 & 0.10 & 1.62
  & 0.36 & 0.05 & 1.70 \\
Sample4Geo~\cite{Sample4geo}
  & 8.80 & 2.29 & 32.55
  & 7.50 & 1.54 & 30.19
  & 1.23 & 0.25 & 5.75
  & 0.50 & 0.14 & 2.11 \\
DetGeo~\cite{DetGeo}
  & 31.18 & 33.15 & 54.08
  & 28.64 & 31.63 & 46.65
  & 46.04 & 55.21 & 59.23
  & 24.13 & 29.04 & 32.20 \\
OCGNet~\cite{OCGNet}
  & 29.51 & 32.28 & 49.18
  & 27.19 & 29.40 & 43.43
  & 47.08 & 56.14 & 60.76
  & 26.77 & 32.52 & 36.19 \\
TROGeo~\cite{TROGeo}
  & 46.86 & 48.67 & 81.27
  & 50.30 & 51.86 & 84.63
  & 51.58 & 60.76 & 69.50
  & 32.66 & 39.22 & 45.46 \\
\midrule
\rowcolor{gray!8}
Ours (Point)
  & 68.48 & 77.01 & 92.86
  & 71.74 & 82.80 & 93.95
  & \textbf{71.50} & 80.70 & \underline{89.69}
  & 65.30 & 72.75 & 81.28 \\
\rowcolor{gray!8}
Ours (Bbox)
  & \underline{74.47} & \underline{85.90} & \underline{97.21}
  & \underline{77.03} & \underline{89.19} & \underline{97.62}
  & \textbf{71.50} & \underline{80.94} & \textbf{89.78}
  & \underline{66.70} & \textbf{74.72} & \underline{82.48} \\
\rowcolor{gray!8}
Ours (Mask)
  & \textbf{76.00} & \textbf{88.32} & \textbf{97.71}
  & \textbf{79.16} & \textbf{91.77} & \textbf{98.26}
  & 71.41 & \textbf{81.04} & 89.39
  & \textbf{67.04} & \textbf{74.72} & \textbf{82.94} \\
\bottomrule
\end{tabular}}
\end{table*}

\begin{table*}[!t]
\centering
\caption{\textbf{Comparison on cross-view object segmentation task.} ``$+$\,SPS'' denotes combination with SAM Prompt Stage~\cite{TROGeo}. The best and second-best results are marked in \textbf{bold} and \underline{underline}, respectively.}
\label{tab:seg_comparison}
\setlength{\tabcolsep}{2pt}
\resizebox{\textwidth}{!}{%
\begin{tabular}{@{} l *{4}{c} @{\hskip 6pt} *{4}{c}  @{\hskip 8pt} *{4}{c} @{\hskip 6pt} *{4}{c} @{}}
\toprule
\multirow{3}{*}{\textbf{Method}}
  & \multicolumn{8}{c}{\textbf{Drone} $\rightarrow$ \textbf{Satellite}}
  & \multicolumn{8}{c}{\textbf{Ground} $\rightarrow$ \textbf{Satellite}} \\
\cmidrule(lr){2-9} \cmidrule(lr){10-17}
  & \multicolumn{4}{c}{Seen} & \multicolumn{4}{c}{Unseen}
  & \multicolumn{4}{c}{Seen} & \multicolumn{4}{c}{Unseen} \\
\cmidrule(lr){2-5} \cmidrule(lr){6-9} \cmidrule(lr){10-13} \cmidrule(lr){14-17}
  & mIoU${\uparrow}$
  & mDice${\uparrow}$
  & AAE${\downarrow}$
  & ME${\downarrow}$
  & mIoU${\uparrow}$
  & mDice${\uparrow}$
  & AAE${\downarrow}$
  & ME${\downarrow}$
  & mIoU${\uparrow}$
  & mDice${\uparrow}$
  & AAE${\downarrow}$
  & ME${\downarrow}$
  & mIoU${\uparrow}$
  & mDice${\uparrow}$
  & AAE${\downarrow}$
  & ME${\downarrow}$ \\
\midrule
Sample4Geo~\cite{Sample4geo}
  & 38.18 & 53.73 & 3407.4 & 18.18
  & 37.61 & 53.05 & 4483.2 & 20.85
  & 23.59 & 35.50 & 14400.8 & 53.87
  & 15.69 & 24.65 & 11538.6 & 89.03 \\
Sample4Geo + SPS
  & 50.09 & 61.58 & 3155.0 & 16.79
  & 53.09 & 63.81 & 4147.4 & 19.16
  & 19.02 & 27.10 & 15946.0 & 55.34
  & 12.15 & 17.81 & 13224.0 & 88.39 \\
DetGeo~\cite{DetGeo}
  & 36.27 & 45.91 & 3882.5 & 55.68
  & 32.20 & 39.42 & 4533.9 & 103.62
  & 33.47 & 42.91 & 15674.4 & 78.32
  & 21.38 & 26.45 & 11738.0 & 147.70 \\
DetGeo + SPS
  & 46.14 & 52.35 & 1786.3 & 55.16
  & 41.83 & 45.67 & 2307.3 & 103.00
  & 48.86 & 53.59 & 3936.1 & 78.57
  & 29.69 & 31.89 & 5661.0 & 147.06 \\
OCGNet~\cite{OCGNet}
  & 33.11 & 42.06 & 4382.4 & 71.85
  & 30.15 & 37.19 & 5192.3 & 105.34
  & 33.76 & 43.39 & 16628.9 & 78.42
  & 23.04 & 28.76 & 13463.0 & 137.24 \\
OCGNet + SPS
  & 43.06 & 48.70 & 1855.4 & 71.52
  & 39.68 & 43.46 & 2520.8 & 104.49
  & 50.07 & 54.91 & 3675.9 & 78.56
  & 32.91 & 35.37 & 5395.1 & 136.26 \\
TROGeo~\cite{TROGeo}
  & 46.86 & 60.69 & 2934.0 & 46.21
  & 49.85 & 63.21 & 3200.2 & 45.01
  & 43.03 & 52.89 & 7452.9 & 115.08
  & 29.80 & 36.98 & 8382.4 & 173.62 \\
TROGeo + SPS
  & 69.09 & 77.99 & 1175.2 & 17.15
  & 75.70 & 82.68 & 1283.6 & 16.90
  & 57.18 & 62.96 & 3291.9 & 59.81
  & 40.71 & 43.95 & 4551.8 & 113.74 \\
\midrule
\rowcolor{gray!8}
Ours (Point)
  & 79.39 & 87.50 & 793.2 & 6.16
  & 81.47 & 88.83 & 810.8 & 5.57
  & \textbf{77.37} & \textbf{84.01} & \underline{1758.1} & \textbf{19.18}
  & 72.68 & 78.63 & 2691.4 & 31.04 \\
\rowcolor{gray!8}
Ours (Bbox)
  & \underline{82.28} & \underline{89.69} & \underline{530.7} & \underline{4.35}
  & \underline{84.12} & \underline{90.86} & \underline{570.1} & \underline{4.02}
  & \underline{77.36} & \underline{83.99} & \textbf{1734.2} & \underline{19.57}
  & \underline{73.82} & \underline{79.76} & \textbf{2528.7} & \underline{29.40} \\
\rowcolor{gray!8}
Ours (Mask)
  & \textbf{83.19} & \textbf{90.35} & \textbf{488.2} & \textbf{4.05}
  & \textbf{85.11} & \textbf{91.58} & \textbf{461.2} & \textbf{3.61}
  & 77.21 & 83.82 & 1770.9 & 19.74
  & \textbf{74.07} & \textbf{80.03} & \underline{2533.4} & \textbf{28.85} \\
\bottomrule
\end{tabular}%
}
\end{table*}

\begin{table*}[t]
\centering
\caption{Zero-shot detection and segmentation performance on CVOGL-Seg datasets (Test set). Note that for segmentation metrics, the comparison methods are equipped with the SAM Prompt Stage (SPS)~\cite{TROGeo}.}
\label{tab:zero_shot_cvogl}
\resizebox{\textwidth}{!}{
\begin{tabular}{@{} l 
*{3}{c} @{\hskip 4pt} *{4}{c} @{\hskip 8pt}
*{3}{c} @{\hskip 4pt} *{4}{c} @{}}
\toprule

\multirow{3}{*}{\textbf{Method}}
& \multicolumn{7}{c}{\textbf{Drone} $\rightarrow$ \textbf{Satellite}}
& \multicolumn{7}{c}{\textbf{Ground} $\rightarrow$ \textbf{Satellite}} \\

\cmidrule(lr){2-8}
\cmidrule(lr){9-15}

& \multicolumn{3}{c}{Detection}
& \multicolumn{4}{c}{Segmentation}
& \multicolumn{3}{c}{Detection}
& \multicolumn{4}{c}{Segmentation} \\

\cmidrule(lr){2-4} \cmidrule(lr){5-8}
\cmidrule(lr){9-11} \cmidrule(lr){12-15}

& mAcc${\uparrow}$ & Acc@75${\uparrow}$ & Acc@50${\uparrow}$
& mIoU${\uparrow}$ & mDice${\uparrow}$ & AAE${\downarrow}$ & ME${\downarrow}$
& mAcc${\uparrow}$ & Acc@75${\uparrow}$ & Acc@50${\uparrow}$
& mIoU${\uparrow}$ & mDice${\uparrow}$ & AAE${\downarrow}$ & ME${\downarrow}$ \\

\midrule

Sample4Geo~\cite{Sample4geo}
& 0.70 & 0.21 & 2.88
& 8.73 & 11.90 & \underline{3554.1} & \underline{112.40}
& 0.42 & 0.10 & 1.75
& 5.10 & 7.14 & \underline{3585.1} & \underline{133.47} \\

DetGeo~\cite{DetGeo}
& 12.09 & 11.41 & 21.79
& 19.14 & 21.71 & 5377.1 & 172.24
& 6.15 & 5.76 & 10.89
& 10.12 & 11.43 & 4809.6 & 208.39 \\

OCGNet~\cite{OCGNet}
& \underline{12.75} & \underline{12.13} & \underline{23.84}
& 20.94 & 23.71 & 6893.0 & 159.25
& \underline{8.77} & \underline{8.74} & \underline{16.14}
& \underline{14.09} & \underline{15.76} & 7073.2 & 180.67 \\

TROGeo~\cite{TROGeo}
& 7.10 & 2.77 & 21.89
& \underline{24.30} & \underline{28.66} & 8823.8 & 137.85
& 3.51 & 1.54 & 9.76
& 10.53 & 12.25 & 8175.8 & 193.37 \\

\midrule
\rowcolor{gray!8}
Ours
& \textbf{29.43} & \textbf{30.55} & \textbf{50.27}
& \textbf{46.79} & \textbf{53.38} & \textbf{2191.3} & \textbf{102.38}
& \textbf{19.37} & \textbf{19.28} & \textbf{35.21}
& \textbf{33.02} & \textbf{38.36} & \textbf{3027.6} & \textbf{113.47} \\

\bottomrule
\end{tabular}}
\end{table*}


\subsection{Ablation Study}
\label{subsec:ablations}
 \begin{figure}[!t]
    \centering
\includegraphics[width=0.99\linewidth]{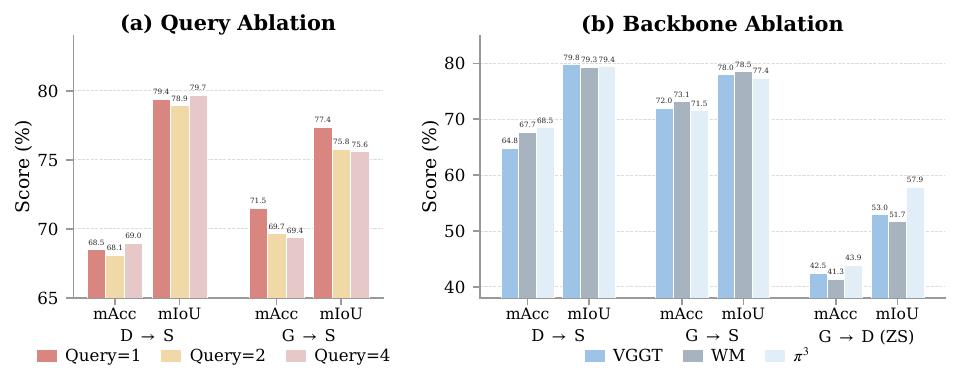}
    \caption{\textbf{Ablation studies of \model.} \textbf{(a) Number of learnable tokens (Query):} Evaluating the impact of query quantity on the D$\rightarrow$S and G$\rightarrow$S setups. \textbf{(b) Backbone architectures:} Comparing our proposed $\pi^3$ backbone against representative 3DFMs (VGGT~\cite{wang2025vggt} and World Mirror~\cite{liu2025worldmirror}). ZS denotes zero-shot.}
    \label{fig:abalation}
\end{figure}

To validate the design choices of \model, we conduct detailed ablation studies on \dataset seen test sets, with all experiments evaluated using point prompts.

\noindent \textbf{Loss ablations.} We evaluate the loss components by treating the grounding loss as the baseline, and incrementally incorporating deep supervision, contrastive learning, and camera pose loss to quantify their cumulative contributions, as shown in Tab.~\ref{tab:ablation_point_seen}.

\begin{wrapfigure}{r}{0.35\textwidth}
  \centering
  \captionof{table}{Zero-shot performance on Ground $\rightarrow$ Drone.}
  \label{tab:zero_shot_g2d}
  \resizebox{0.3\textwidth}{!}{ 
  \small
  \begin{tabular}{l|ccc}
  \toprule
  Method
  & mAcc$\uparrow$
  & mIoU$\uparrow$\\
  \midrule
  Sample4Geo~\cite{Sample4geo} & 0.43 & 12.07\\
  DetGeo~\cite{DetGeo} & 12.60 & 22.83\\
  OCGNet~\cite{OCGNet} & 15.16 & 27.16\\
  TROGeo~\cite{TROGeo} & 20.92 & 35.81\\
  \midrule
  Ours (w/o CL) & 38.38 & 57.00\\
  \rowcolor{gray!8}
  Ours (Point) & \textbf{43.86} & \textbf{57.88}\\
  \bottomrule
  \end{tabular}
  }
\end{wrapfigure}
Notably, Tab.~\ref{tab:zero_shot_g2d} confirms that our contrastive learning loss narrows the ground-to-drone feature gap by using satellite object embeddings as anchors.

\noindent \textbf{Number of learnable tokens.} 
Fig.~\ref{fig:abalation}(a) shows that a single task token is sufficient to capture the target information. Increasing queries to 2 or 4 degrades performance in $G \to S$ without consistent gains in $D \to S$, as additional tokens introduce redundant noise for single-object localization.

\noindent \textbf{Impact of different backbones.} 
As shown in Fig.~\ref{fig:abalation}(b), despite its lighter architecture (18 \vs 24 layers), $\pi^3$ delivers competitive performance and superior zero-shot G$\rightarrow$D generalization. This suggests that the inherited geometric prior and the permutation-equivariant design help reduce view-specific biases.


\begin{table}[!t]
\centering
\caption{Ablation study (seen test set) with \textbf{point} prompts. 
Each row cumulatively adds the indicated module. 
\colorbox{gray!15}{Gray} denotes the full model.}
\label{tab:ablation_point_seen}
\resizebox{\textwidth}{!}{%
\setlength{\tabcolsep}{3pt}
\begin{tabular}{clccccccc}
\toprule
\textbf{Task} & \textbf{Method} & mAcc$\uparrow$ & Acc@50$\uparrow$ & Acc@75$\uparrow$ & mIoU$\uparrow$ & mDice$\uparrow$ & AAE$\downarrow$ & ME$\downarrow$ \\
\midrule
\multirow{4}{*}{D $\rightarrow$ S}
 & (a)~Base                       & 62.86 & 89.97 & 71.06 & 77.19 & 85.82 & 861.1 & 7.22 \\
 & (b)~+ Deep Supervision         & 66.38 & 91.85 & 75.64 & 78.71 & 86.96 & 797.3 & 6.39 \\
 & (c)~+ Contrastive              & 67.63 & 92.11 & 76.92 & \textbf{79.51} & \textbf{87.55} & \textbf{738.9} & 6.18 \\
 & \cellcolor{gray!15}(d)~+ Camera Pose 
 & \cellcolor{gray!15}\textbf{68.48}
 & \cellcolor{gray!15}\textbf{92.86}
 & \cellcolor{gray!15}\textbf{77.01}
 & \cellcolor{gray!15}79.39
 & \cellcolor{gray!15}87.50
 & \cellcolor{gray!15}793.2
 & \cellcolor{gray!15}\textbf{6.16} \\
\midrule
\multirow{4}{*}{G $\rightarrow$ S}
 & (a)~Base                       & 67.41 & 86.64 & 77.01 & 74.86 & 81.78 & 2099.4 & 22.23 \\
 & (b)~+ Deep Supervision         & 70.06 & 88.41 & 79.57 & 76.58 & 83.29 & 1780.0 & 19.54 \\
 & (c)~+ Contrastive              & 71.33 & 89.47 & 80.69 & \textbf{77.50} & \textbf{84.12} & \textbf{1748.8} & 19.47 \\
 & \cellcolor{gray!15}(d)~+ Camera Pose 
 & \cellcolor{gray!15}\textbf{71.50}
 & \cellcolor{gray!15}\textbf{89.69}
 & \cellcolor{gray!15}\textbf{80.70}
 & \cellcolor{gray!15}77.37
 & \cellcolor{gray!15}84.01
 & \cellcolor{gray!15}1758.1
 & \cellcolor{gray!15}\textbf{19.18} \\
\bottomrule
\end{tabular}%
}
\end{table}

\subsection{Qualitative Results}
\label{subsec:qualitative_results}
\begin{figure}[t]
    \centering
\includegraphics[width=0.99\linewidth]{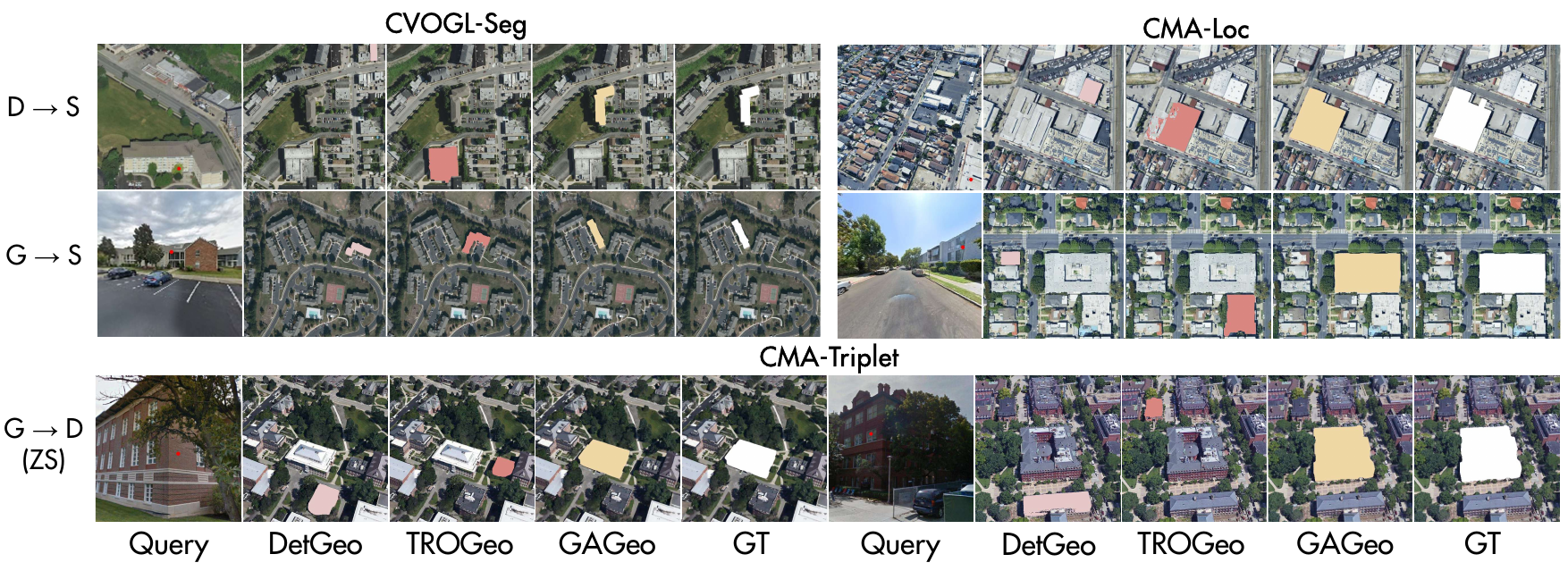}
    \caption{Qualitative Results on CVOGL-Seg~\cite{TROGeo} and \dataset datasets.}
    \label{fig:Qualitative Results}
 \vspace{-2mm}
\end{figure}

Fig.~\ref{fig:Qualitative Results} visualizes the performance across three cross-view setups on CVOGL-Seg and \dataset dataset. Our method yields precise segmentation masks aligned with the ground truth, even under the unseen ground-to-drone setup. In contrast, existing methods like TROGeo~\cite{TROGeo} and DetGeo~\cite{DetGeo} frequently yield fragmented or misaligned predictions, further validating the robustness of \model.

\section{Conclusion}
In this paper, we address the intertwined data and methodological bottlenecks in CVOGL through a comprehensive framework. We introduce \dataset, a large-scale, high-fidelity building dataset that overcomes prior limitations by providing diverse referring prompts and camera metadata across standard field-of-view imagery. To bridge the extreme spatial gap, we propose \model, a unified, geometry-aware architecture that adapts the 3D prior of $\pi^3$ to simultaneously output multi-task predictions in a single streamlined forward pass. Furthermore, by utilizing the satellite view as a universal bridge in our contrastive learning objective, \model enhances zero-shot ground-to-drone localization without relying on scarce triplet data. Extensive experiments demonstrate that our approach establishes a new state-of-the-art, advancing both detection and segmentation performance while exhibiting robust cross-domain generalization.

Admittedly, the inference efficiency of our framework is currently limited by the architectural complexity of 3D foundation models. Promising avenues for future research involve leveraging established paradigms such as token merging, pruning, and knowledge distillation to mitigate these efficiency constraints.

\section*{Acknowledgements}
This research is supported in part by the Key Research Program of Hangzhou (No. 2025SZD1A56), the National Natural Science Foundation of China (No. 62461160308, U23B2010, 62576024), the Beijing Natural Science Foundation (No. L231011), the Fundamental Research Funds for the Central Universities (No. 501RCQD2025141003), BeiHang GanWei Project (No. 502GWXM20241410\allowbreak01), the National Science Foundation Support Projects (No. 62425303), the National Key R\&D Program of China (No. 2024YFB4707300), and the Beijing Nova Program.
%
%
\bibliographystyle{splncs04}
\bibliography{main}
\end{document}